\documentclass[acmtog,screen,nonacm,author-version]{acmart}
\AtBeginDocument{%
  }

\setcopyright{acmlicensed}
\copyrightyear{2018}
\acmYear{2018}
\acmDOI{XXXXXXX.XXXXXXX}
\acmConference[Conference acronym 'XX]{Make sure to enter the correct
  conference title from your rights confirmation email}{June 03--05,
  2018}{Woodstock, NY}
\acmISBN{978-1-4503-XXXX-X/2018/06}

\begin{document}

\title{Suzume-chan: Your Personal Navigator as an Embodied Information Hub}

\author{Maya Grace Torii}
\email{toriparu@digitalnature.slis.tsukuba.ac.jp}
\affiliation{%
  \institution{Doctoral Program in Informatics, University of Tsukuba}
  \city{Tsukuba}
  \state{Ibaraki}
  \country{Japan}
}

\author{Takahito Murakami}
\email{takahito@digitalnature.slis.tsukuba.ac.jp}
\affiliation{%
  \institution{Doctoral Program in Informatics, University of Tsukuba}
  \city{Tsukuba}
  \state{Ibaraki}
  \country{Japan}
}

\author{Shuka Koseki}
\email{s2430442@u.tsukuba.ac.jp}
\affiliation{%
  \institution{Doctoral Program in Nursing Science, University of Tsukuba}
  \city{Tsukuba}
  \state{Ibaraki}
  \country{Japan}
}

\author{Yoichi Ochiai}
\affiliation{%
  \institution{Institute of Library, Information and Media Science, University of Tsukuba}
  \city{Tsukuba}
  \state{Ibaraki}
  \country{Japan}
}
\email{wizard@slis.tsukuba.ac.jp}

\renewcommand{\shortauthors}{Trovato et al.}

\begin{abstract}
Access to expert knowledge often requires real-time human communication.  
Digital tools improve access to information but rarely create the sense of connection needed for deep understanding.
This study addresses this issue using Social Presence Theory, which explains how a feeling of “being together” enhances communication.
An “Embodied Information Hub” is proposed as a new way to share knowledge through physical and conversational interaction.
The prototype, Suzume-chan, is a small, soft AI agent running locally with a language model and retrieval-augmented generation (RAG).
It learns from spoken explanations and responds through dialogue, reducing psychological distance and making knowledge sharing warmer and more human-centered.
\end{abstract}



\begin{teaserfigure}
  \includegraphics[width=\textwidth]{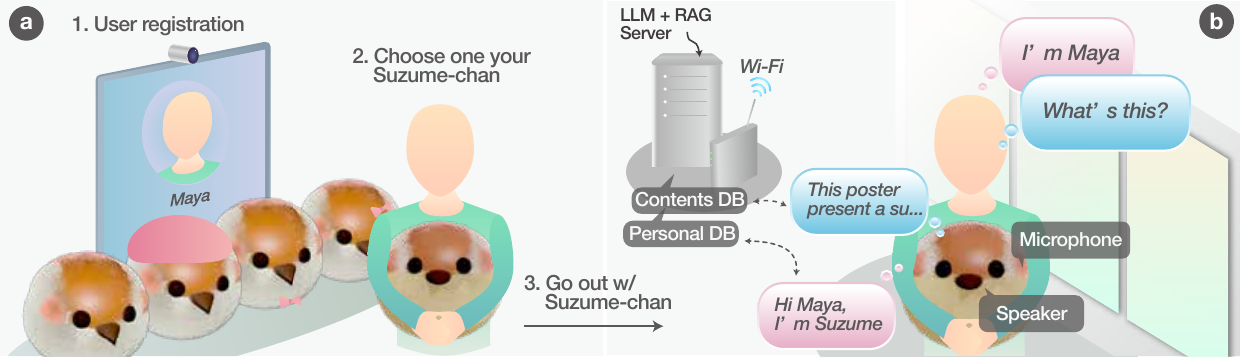}
  \caption{Overview of the demonstration. (a) On-site user experiences, and (b) system overview}
  \label{fig:teaser}
\end{teaserfigure}

\maketitle

\section{Introduction}
In the modern era, people seek not only data but also the stories that lie behind knowledge—such as the passion of experts or the background of artworks. However, access to such stories has often required direct and synchronous communication with experts. Smartphone applications provide efficient access to information, but they tend to isolate users from their surroundings. They can deliver answers but do not create relationships.

This study examines the absence of such relationships. The quality of human communication has been described by Social Presence Theory \cite{short1976social}, which explains how the sense that “someone is there” enriches mediated interaction. We ask what happens when this human-centered theory is applied to an AI agent with a physical body. This is the main question of our research.

Previous studies such as Paro \cite{shibata2001mental}, Kismet \cite{breazeal2004designing}, and iCub \cite{metta2008icub} have shown that physical robots can form emotional and social bonds with humans. These studies suggest that an agent can become a social partner rather than a mere tool.

We extend this possibility from the emotional domain to the intellectual one. Suzume-chan was intentionally designed with a soft and friendly appearance to explore this new type of relationship. Its warm, hand-sized form helps reduce psychological barriers and turns intellectual explanation into a calm and trustworthy conversation. The “Embodied Information Hub” proposed in this paper aims to explore a new form of social presence between humans and AI agents, and Suzume-chan represents its first implementation.

\section{Method: Suzume-chan}
\subsection{System Overview}
Suzume-chan consists of two components: a handheld agent used by visitors for conversation, and a host computer that performs high-load language processing. These two units are connected wirelessly via Wi-Fi, allowing the system to operate in a fully standalone environment.

\subsubsection{Hardware}
The agent body is covered with a soft plush exterior to provide a sense of psychological safety. Inside the body, a microphone captures the speaker’s voice and a speaker outputs Suzume-chan’s responses. The host computer is a Mac Studio equipped with 128 GB of unified memory. It runs open-source models locally, including a speech recognition model (e.g., Whisper\cite{radford2023robust}), large language models (LLM;  e.g., Llama\cite{touvron2023llama}, gpt-oss-120b\cite{agarwal2025gpt}), a vector database\cite{topsakal2023creating}, and a speech synthesis engine. This configuration ensures both privacy protection and stable operation without dependence on an external network.

\subsection{Software}
The dialogue engine is based on a Retrieval-Augmented Generation (RAG) framework\cite{lewis2020retrieval}.
During the input phase, the spoken explanations are transcribed and converted into vector representations, which are stored in the database.
During the explanation phase, visitors’ questions are also vectorized, and the system searches for relevant information within the database. The retrieved results are included in the prompt for the language model, which generates natural and contextually accurate responses.

The interaction with Suzume-chan follows a simple two-phase process. In the input phase, the presenter explains their research content to Suzume-chan, which processes the spoken information, divides it into smaller chunks, and stores them as vector representations in the database. In the explanation phase, visitors can initiate a conversation by using a wake word (e.g., “Hey, Suzume-chan”) (Fig.\ref{fig:teaser}(b)).
When they ask questions such as “What is special about this research?”, Suzume-chan retrieves relevant information from the database, summarizes it, and provides a natural and accessible response.

\section{Originality and Contributions}
The originality of this study lies in proposing the concept of a “Physical Information Hub,” which expands the role of physical agents from providing emotional comfort to mediating expert knowledge. Conceptually, this work introduces a new type of agent that enables asynchronous mediation of expert knowledge. Technically, it implements a standalone dialogue system that combines a local LLM and RAG framework to ensure both privacy and usability. Practically, it presents a working prototype that addresses real-world challenges in academic knowledge sharing.

\section{Demonstration}
We conduct an empirical study with visitors in WISS 2025, Japan\cite{WISS2025}. Before the session, presenters will teach Suzume-chan about their own research topics and this project. Visitors will freely interact with the system, and their interactions will be observed. After the experience, semi-structured interviews and questionnaires will be conducted through Suzume-chan. Through these procedures, we will evaluate the system’s usefulness, acceptability, and interaction design issues, and collect initial evidence regarding the effectiveness of Suzume-chan (Fig.\ref{fig:teaser}(a)).

\section{Conclusion}
This study proposed the “Physical Information Hub” as a new paradigm for sharing expert knowledge beyond temporal and spatial constraints. The prototype, Suzume-chan, functions as a standalone agent that combines local LLM and RAG technologies to mediate knowledge asynchronously and interactively. Its system design and evaluation plan demonstrate the potential for wider application and suggest a step toward technology that fosters warm, human-centered communication.

\begin{acks}
This research was supported by the Young Researcher Support Program of the Center for Cyber Medicine Research, University of Tsukuba.
\end{acks}

\bibliographystyle{ACM-Reference-Format}
\bibliography{sample-base}

\newpage
\appendix
\section{Future vision}
\subsection{1. Input and Explanation Phases — One-to-One Relationship Between Human and Agent:}
Suzume-chan learns, remembers, and explains within an individual relationship.
Users share knowledge and experiences, which are vectorized and stored.
The agent replies using past context, creating continuity and trust.
Its gradual growth embodies a new human–AI relationship.

\subsection{2. Conversational Survey — A New Form of Data Collection:}
Agents can collect information naturally through dialogue instead of forms.
Suzume-chan’s questions record contextual, qualitative data, making surveys more engaging and less burdensome.
This conversational method offers a new way to gather insights in research, education, and public design.

\subsection{3. Suzume Network — Sharing Knowledge Among Agents:}
In the “Suzume Network,” agents share user-consented experiences like word of mouth.
Individual interactions become collective knowledge, enriching context and continuity.
Through \textit{learning}, \textit{listening}, and \textit{sharing}, Suzume-chan evolves into a living hub for collective human–AI memory.

\end{document}